\documentclass[10pt,twocolumn,letterpaper]{article}

\usepackage{cvpr}
\usepackage{times}
\usepackage{epsfig}
\usepackage{graphicx}
\usepackage{amsmath}
\usepackage{amssymb}
\usepackage{multirow}
\usepackage{diagbox}
\usepackage{tabularx}
\usepackage{caption}
\usepackage[symbol]{footmisc}
\usepackage[page]{appendix}

\usepackage[breaklinks=true,bookmarks=false]{hyperref}

\newcommand{\tabincell}[2]{\begin{tabular}{@{}#1@{}}#2\end{tabular}}  

\cvprfinalcopy 

\def\cvprPaperID{5384} 

\ifcvprfinal\fi
\begin{document}

\title{Towards High-Fidelity 3D Face Reconstruction from In-the-Wild Images Using Graph Convolutional Networks}

\author{Jiangke Lin, Yi Yuan\footnotemark\\
NetEase Fuxi AI Lab\\
Hangzhou, China\\
{\tt\small \{linjiangke, yuanyi\}@corp.netease.com}
\and
Tianjia Shao, Kun Zhou\\
State Key Lab of CAD\&CG, Zhejiang University\\
Hangzhou, China\\
{\tt\small tianjiashao@gmail.com, kunzhou@acm.org}
}

\twocolumn[{
\renewcommand\twocolumn[1][]{#1}
\maketitle
\thispagestyle{empty}
\begin{center}
   \centering
   \includegraphics[width=0.9\textwidth]{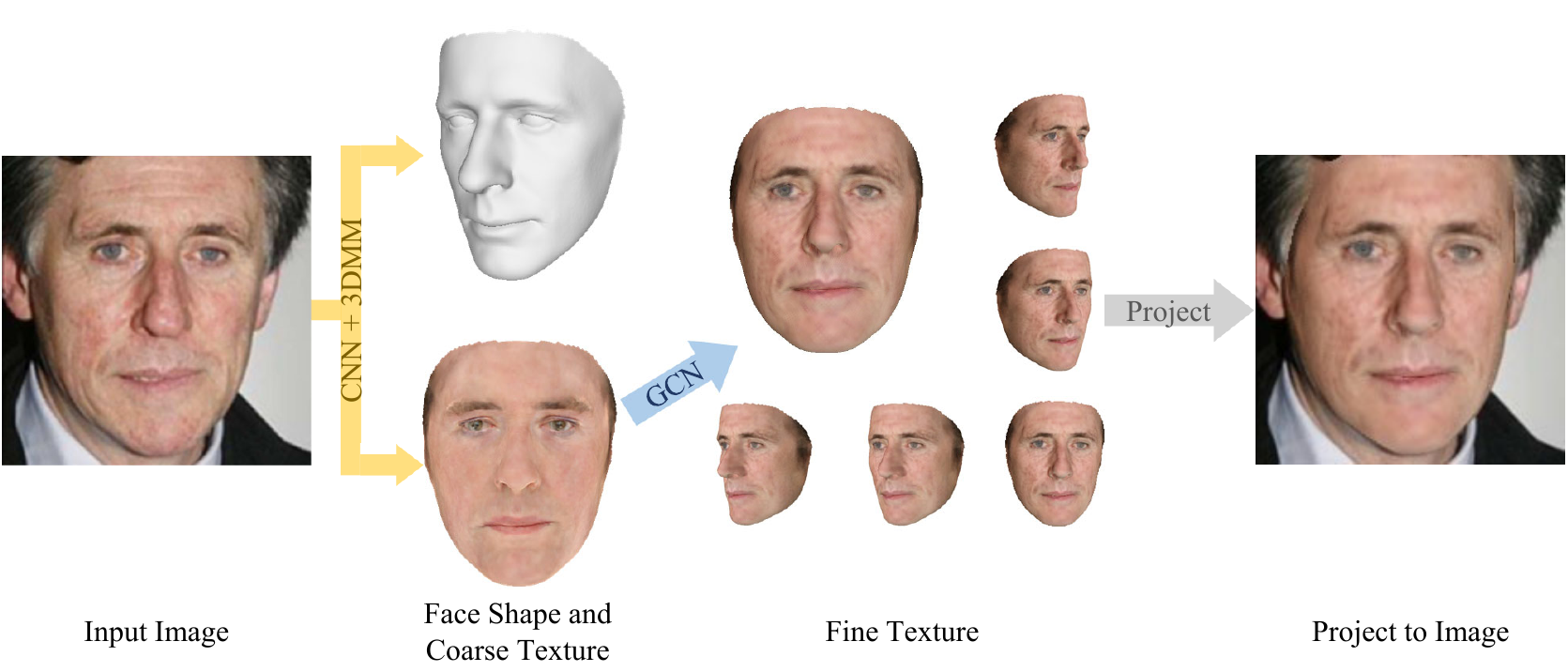}
   \captionof{figure}{From left to right: the input image, the reconstructed face shape and coarse texture, the fine detailed texture, and the 3D face projected to the input image.}
\end{center}
}]

\footnotetext[1]{Corresponding author}
\footnotetext[2]{The code is available at https://github.com/FuxiCV/3D-Face-GCNs}
\begin{abstract}
3D Morphable Model (3DMM) based methods have achieved great success in recovering 3D face shapes from single-view images. However, the facial textures recovered by such methods lack the fidelity as exhibited in the input images. Recent work demonstrates high-quality facial texture recovering with generative networks trained from a large-scale database of high-resolution UV maps of face textures, which is hard to prepare and not publicly available. In this paper, we introduce a method to reconstruct 3D facial shapes with high-fidelity textures from single-view images in-the-wild, without the need to capture a large-scale face texture database. The main idea is to refine the initial texture generated by a 3DMM based method with facial details from the input image. To this end, we propose to use graph convolutional networks to reconstruct the detailed colors for the mesh vertices instead of reconstructing the UV map. Experiments show that our method can generate high-quality results and outperforms state-of-the-art methods in both qualitative and quantitative comparisons.
\end{abstract}
\vspace*{-6mm}

\section{Introduction}
Reconstructing the 3D facial shape and texture from a single image is a vital problem in computer vision and graphics. The seminal work of Blanz and Vetter~\cite{blanz1999morphable} illustrates the power of the 3D Morphable Model (3DMM), which is a statistical model of facial shape and texture built from hundreds of scanned faces. The 3DMM and its variants~\cite{booth2018large, cao2013facewarehouse, gerig2018morphable, huber2016multiresolution, li2017learning} make it possible to recover the facial shape and albedo from a single image~\cite{deng2019accurate, genova2018unsupervised, sanyal2019learning, wu2019mvf}. However, the fidelity of the texture recovered from the 3DMM coefficients is still not high enough. It is mainly because that the texture computed from 3DMM cannot capture the face details of the input image, especially for the images in-the-wild.


In order to tackle the problem, recently there have been some work~\cite{deng2018uv, gecer2019ganfit} trying to reconstruct high-quality textures from 2D images. For example, Deng \etal.~\cite{deng2018uv} learn a generative model to complete the self-occluded regions in the facial UV map. Gecer \etal.~\cite{gecer2019ganfit} first utilize GANs to train a generator of facial textures in UV space and then use non-linear optimization to find the optimal latent parameters. While they can achieve high fidelity textures, their methods require a large-scale database of high-resolution UV maps, which is not publicly available. Besides, capturing such a database is rather laborious, which is infeasible for ordinary users.

In this paper, we seek to reconstruct the 3D facial shape with high fidelity texture from a single image, without the need to capture a large-scale face texture database. To achieve this goal, we make a key observation that, though lacking details, the 3DMM texture model can provide a globally-reasonable color for the whole face mesh. We can further refine this initial texture by introducing the facial details from the image to the face mesh. To this end, we propose to reconstruct the detailed colors for the mesh vertices instead of reconstructing the UV map. In particular, we utilize graph convolutional networks (GCN)~\cite{defferrard2016convolutional, kipf2016semi, ranjan2018generating} to decode the image features and propagate the detailed RGB values to the vertices of the face mesh.

Our reconstruction framework works in a coarse-to-fine manner, based on a combination of a 3DMM model and GCNs. A convolutional neural network (CNN) is trained to regress 3DMM coefficients (identity/expression/texture) and rendering parameters (pose/lighting) from a 2D image. With the 3DMM model, the face shape and initial coarse texture can be computed with an affine model. At a key stage, we utilize a pre-trained CNN to extract face features from the image and feed them to a graph convolutional network to produce detailed colors for the mesh vertices. Our framework adopts a differentiable rendering layer~\cite{genova2018unsupervised} to enable self-supervised training, and further improves the results with a GAN loss~\cite{gulrajani2017improved}.

To summarize, this paper makes the following contributions:
\begin{itemize}
\item We propose a coarse-to-fine framework for reconstructing 3D faces with high fidelity textures from single images, without requiring to capture a large scale of high-resolution face texture data.
\item To the best of our knowledge, we are the first to use graph convolutional networks to generate high fidelity face texture, which is able to generate detailed colors for the mesh vertices from the image.
\item We compare our results with state-of-the-art methods, and ours outperform others in both qualitative and quantitative comparison.
\end{itemize}

\begin{figure*}
\begin{center}
\includegraphics[width=0.9\linewidth]{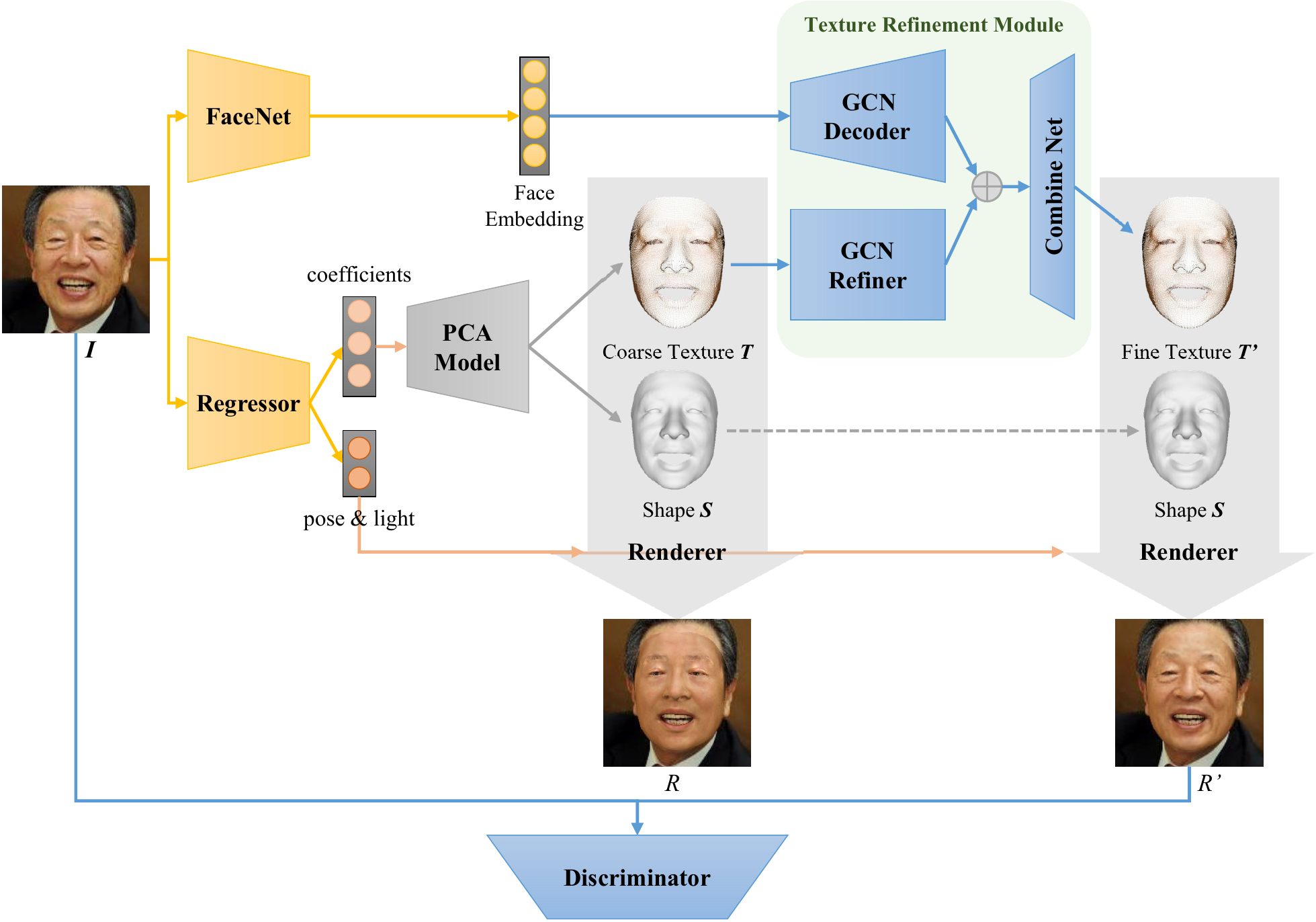}
\end{center}
   \caption{The overall coarse-to-fine framework of our approach. The yellow blocks are pre-trained before others, the grays are non-trainable and the blues are trainable. The Regressor regresses the 3DMM coefficients, face pose and lighting parameters from the input image $I$, where the 3DMM coefficients will be used to compute the face shape $S$ (coordinates, \eg $x, y, z$) and coarse texture $T$ (albedo, \eg $r, g, b$) through the PCA model. The FaceNet is used to extract a face embedding from $I$. Then the coarse texture and the face embedding are fed into the GCN Refiner and GCN Decoder respectively. The outputs of the two GCNs are concatenated along the channel axis (donated as $\oplus$) and fed to the Combine Net, which produces the fine texture $T'$. The Discriminator attempts to improve the output of texture refinement module via adversarial training.}
\label{fig:overall}
\end{figure*}

\section{Related Work}
\subsection{Morphable 3D Face Models}
Blanz and Vetter~\cite{blanz1999morphable} introduced the first 3D morphable face model twenty years ago. Since then, there have been multiple variations of 3DMM~\cite{booth2018large, cao2013facewarehouse, gerig2018morphable, huber2016multiresolution, li2017learning}. Those models produce low-dimensional representations for the facial identity, expression and texture from multiple face scans using PCA. One of the most widely used, publicly available variants of 3DMM is the Basel Face Model (BFM)~\cite{paysan20093d}. It registers a template mesh to the scanned faces with the Optimal Step Nonrigid ICP algorithm, then employs PCA for dimensionality reduction to construct the model. We will use this model as our 3DMM model in our experiments.

In a 3DMM, given the identity coefficients $c_{i}$, expression coefficients $c_{e}$ and texture coefficients $c_{t}$, the face shape $S$ and the texture $T$ can be represented as:
\begin{equation}
\begin{aligned}
   &S = S_{mean} + c_{i}I_{base} + c_{e}E_{base}\\
   &T = T_{mean} + c_{t}T_{base}
\end{aligned}
\label{con:3DMM}
\end{equation}
where $S_{mean}$ and $T_{mean}$ are the mean face shape and texture, and $I_{base}$, $E_{base}$ and $T_{base}$ are the PCA bases of identity, expression and texture respectively. We use $S_{mean}$, $T_{mean}$, $I_{base}$ and $T_{base}$ from BFM~\cite{paysan20093d}, and $E_{base}$ built from FaceWarehouse~\cite{cao2013facewarehouse}.

\subsection{Fitting a Morphable Face Model}
A classical approach for 3D face reconstruction from a single image is to iteratively fit a template model to the input 2D image. However, this approach is sensitive to the lighting, expression, and pose of the 2D face image. Some improvements~\cite{blanz2003face, levine2009state, romdhani2005estimating} have been made to improve the fitting stability and accuracy, but they don't perform well on images in the wild.

Deep learning based methods directly regress the 3DMM coefficients from images. To obtain paired 2D-3D data for supervised learning, Richardson \etal ~\cite{richardson20163d, richardson2017learning} generate synthetic data by random sampling from the morphable face model, but this approach may not perform well when dealing with complex lighting, occlusion, and other in-the-wild conditions. Tran \etal \cite{tuan2017regressing} do not generate synthetic data directly. Instead, they create the ground truth using an iterative optimization to fit a large number of face images. Nevertheless, the problem of being delicate in uncontrolled environments still remains.

Recently, differentiable renderers~\cite{genova2018unsupervised, tewari2017mofa} has been introduced. It renders a 3D face mesh to a 2D image based on the face shape, texture, lighting and other related parameters, and compare the rendered image with the input image to compute the loss in terms of image differences. With such a fixed, differentiable rendering layer, unsupervised or weakly-supervised training is enabled without requiring the training pairs. We also follow this strategy. Specially, we adopt the differentiable renderer from Genova \etal~\cite{genova2018unsupervised}, which is known as "tf-mesh-renderer", as our differentiable rendering layer.

\subsection{Texture of Morphable Face Models}
Recent deep learning based methods aim to reconstruct the facial textures from single images by regressing the 3DMM texture coefficients. For example, Deng \etal~\cite{deng2019accurate} proposed a method to simultaneously predict the 3DMM shape and texture coefficients, which employs an illumination and rendering model during training, and conducts image-level and perception-level losses, leading to a better result than others. However, the texture they generated is still inherently limited by the 3DMM texture model. In our method, we deploy their scheme to train a 3DMM coefficient regression model, to get a globally reasonable face texture. Then we utilize graph convolutional networks to refine the texture with the image details.

\subsection{Graph Convolutional Networks}
To enable convolutional operations in non-Euclidean structured data, Bruna \etal~\cite{bruna2013spectral} utilize the graph Laplacian and the Fourier basis to make the first extension of CNNs on graphs. However, it is computationally expensive and ignores the local features. Defferrard \etal~\cite{defferrard2016convolutional} propose the ChebyNet, which approximates the spectral filters by truncated Cehbyshev polynomials, avoiding the computation of the Fourier basis. CoMA~\cite{ranjan2018generating} introduces mesh downsampling and mesh upsampling layers, which constructs an autoencoder to learn a latent embedding of 3D face meshes. Motivated by CoMA~\cite{ranjan2018generating}, our method learns a latent vector representing the detailed face color from a 2D image, then decodes it to produce detailed colors for the face mesh vertices with graph convolutional networks.

\section{Approach}
We propose a coarse-to-fine approach for 3D face reconstruction. As shown in Fig.~\ref{fig:overall}, our framework is composed of three modules. The feature extraction module includes a Regressor for regressing the 3DMM coefficients, face pose, and lighting parameters, and a FaceNet~\cite{schroff2015facenet} for extracting image features for the subsequent detail refinement and identity-preserving. The texture refinement module consists of three graph convolutional networks: a GCN Decoder to decode the features extracted from FaceNet and producing detailed colors for mesh vertices, a GCN Refiner to refine the vertex colors generated from the Regressor, and a combiner to combine the two colors to produce final vertex colors. The Discriminator attempts to improve the output of the texture refinement module via adversarial training.

\subsection{3DMM Coefficients Regression}
The first step of our algorithm is to regress the 3DMM coefficients and rendering parameters from the input image with CNNs. We adopt the state-of-the-art 3DMM coefficient regressor in~\cite{deng2019accurate} for this task. Given a 2D image $I$, it regresses a 257 dimensional vector $(c_{i}, c_{e}, c_{t}, p, l)\in\mathbb{R}^{257}$, where $c_{i}\in\mathbb{R}^{80}$, $c_{e}\in\mathbb{R}^{64}$ and $c_{t}\in\mathbb{R}^{80}$ represent the 3DMM identity, expression and texture coefficients respectively. $p\in\mathbb{R}^{6}$ is the face pose and $l\in\mathbb{R}^{27}$ is lightings. With the predicted coefficients, the face vertices' 3D positions $S$ and albedo values $T$ can be computed with Eq.~\ref{con:3DMM}.

Moreover, we utilize a pre-trained FaceNet~\cite{schroff2015facenet} to extract a feature vector from the face image. The extracted feature serves two purposes. First, it can be treated as an image feature embedding for our Decoder to generate detailed albedo colors for the mesh vertices, which is described in Sec. \ref{sec:refine}. Second, it can be used to measure the identity distance in the identity-preserving loss in Sec. \ref{sec:losses}.

\subsection{Differentiable Rendering}\label{sec:render}
In order to train our networks, we conduct a self-supervised approach with a differentiable rendering layer. That is, we render the face mesh to a 2D image with the predicted parameters, and compute the losses based on the differences between the rendered image and the input image. We adopt the differentiable rendering layer from~\cite{genova2018unsupervised}, which introduced a general-purpose, differentiable rasterizer based on a deferred shading model. The rasterizer computes screen-space buffers with triangle IDs and barycentric coordinates for the pixels. The colors and normals from the mesh are interpolated at the pixels. During training, the vertex normals is computed as the average of surrounding triangle normals.

\begin{figure}[h]
\begin{center}
\includegraphics[width=0.95\linewidth]{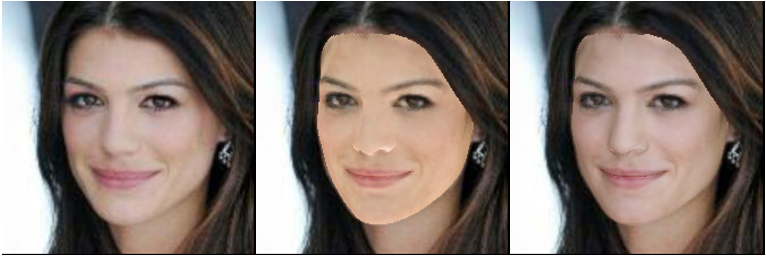}
\end{center}
   \caption{From left to right: the input image, the rendered image with only albedo, and the rendered image with illuminated texture.}
\label{fig:albedo}
\end{figure}

Specifically, with the shape $S$, texture $T$ and pose generated from the Regressor, we can compute the face albedo projected on the input image. As the projected face albedo is not the final result, we further illuminate the face mesh with the estimated lighting and render it to get the final image $R$, which is compared with the input image to compute the loss. An example of illumination and rendering is shown in Fig.~\ref{fig:albedo}.

\subsection{Texture Refinement Module}\label{sec:refine}
Our texture refinement module is composed of three graph convolutional networks, namely a Decoder, a Refiner and a combine net. Unlike other work~\cite{deng2018uv, gecer2019ganfit} using the UV map as the face texture representation, we directly manipulate the albedo RGB values of the vertices on the face mesh. We deploy a face texture mesh consisting of a set of vertices and triangles, denoted as $\mathcal{M=(V, A)}$, where $\mathcal{V} \in \mathbb{R}^{n \times 3}$ stores vertex colors, and $\mathcal{A} \in \{0, 1\}^{n \times n}$ is the adjacency matrix, representing the triangles. In the adjacency matrix $\mathcal{A}$, if vertex $i$ and $j$ are connected, then $\mathcal{A}_{ij} = 1$, and $\mathcal{A}_{ij} = 0$ otherwise. For normalization, a Laplacian matrix is computed as $\mathcal{L=I}-D^{-\frac{1}{2}}\mathcal{AD}^{-\frac{1}{2}}$, where $\mathcal{I}$ is the identity matrix, and $\mathcal{D}$ is the diagonal matrix representing the degree of each vertex in $\mathcal{V}$ as $\mathcal{D}^{ii} = \sum_j \mathcal{A}^{ij}$. The spectral graph convolution operator * between $x$ and $y$ is defined as a Hadamard product in the Fourier space: $x * y = U((U^T x) \odot (U^T y))$, where $U$ is the eigenvectors of Laplacian matrix. To address the problem of computationally expensive caused by non-sparsity of $U$, \cite{defferrard2016convolutional} proposed a fast spectral convolution method, which constructs mesh filtering with a kernel $g_\theta$ using a recursive Chebyshev polynomial~\cite{defferrard2016convolutional, hammond2011wavelets}:
\begin{equation}
   g_\theta(\mathcal{L}) = \sum_{k=0}^{K-1} \theta_k T_k (\mathcal{\widetilde{L}}),
\label{con:cheby}
\end{equation}
where $\mathcal{\widetilde{L}} = 2 \mathcal{L} / \lambda_{max} - I$ is the scaled Laplacian matrix, $\lambda_{max}$ is the maximum eigenvalue of the Laplacian matrix, $\theta \in \mathbb{R}^K$ is the Chebyshev coefficients vector, and $T_k \in \mathbb{R}^{n \times n}$ is the Chebyshev polynomial of order $k$. $T_k$ is computed recursively as $T_k(x) = 2x T_{k-1}(x) - T_{k-2}(x)$ with the initial $T_0 = 1$ and $T_1 = x$. For each layer, the spectral convolution can be defined as:
\begin{equation}
   y_j = \sum_{i=1}^{F_in} g_{\theta_{i,j}} (\mathcal{L}) x_i \in \mathbb{R}^{n},
\label{con:spectral_conv}
\end{equation}
where $x \in \mathbb{R}^{n \times F_{in}}$ is the input with $F_{in}$ features, and $y_j$ denote the $j^{th}$ feature of $y \in \mathbb{R}^{n \times F_{out}}$, which has $F_{out}$ features. For each spectral convolution layer, there are $F_{in} \times F_{out}$ trainable parameters.

Our Decoder takes as input the feature vector from FaceNet and produces the albedo RGB values for each vertex. The Decoder architecture is built following the idea of residual networks, which consists of four spectral residual blocks. The spectral upsampling layers are placed between each two spectral residual blocks. Each spectral residual block contains two Chebyshev convolutional layers and one shortcut layer. Every Chebyshev convolutional layer uses $K=6$ Chebyshev polynomials and is followed by a biased ReLU layer~\cite{glorot2011deep}.

The Refiner refines the vertex colors from the 3DMM model with spectral convolutional layers. It also contains spectral residual blocks similar to the Decoder. However, only one downsampling layer and one upsampling layer are deployed in the bottom and at the top of the network separately.

To produce the final vertex colors with details, the combination net concatenates the outputs of the Decoder and Refiner along the channel axis, and feed them into a graph convolutional layer, followed by a \textit{tanh} activation layer.

We also adopt the same differentiable renderer to self-supervise the training of our texture refinement module as in Sec.~\ref{sec:render}. Specifically, we render the 3D face mesh to $R'$ with illumination and compare it with the input image. To make the final texture own higher fidelity, we further utilize an adversarial training strategy. Since we do not use the real 3D face data, the Discriminator is deployed on the input 2D images and the rendered images from the reconstructed 3D face meshes. Our Discriminator contains 6 convolutional layers with kernel size 3, each of which is followed by a max-pooling layer. During training, we follow the procedure described in Wasserstein GANs with gradient penalty~\cite{gulrajani2017improved}.

\begin{figure}
\begin{center}
\includegraphics[width=0.8\linewidth]{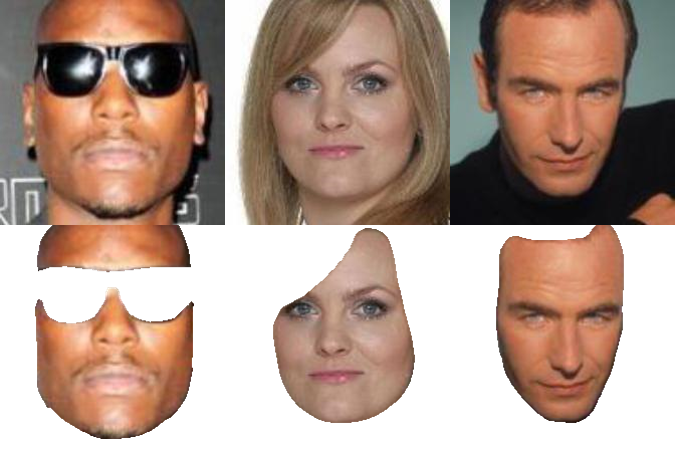}
\end{center}
   \caption{The first row is the input images, and the second row is the face regions for computing losses.}
\label{fig:skin_region}
\end{figure}

\subsection{Losses}\label{sec:losses}
\subsubsection{Pixel-wise Loss}
A straightforward objective is to minimize the differences between the input images and the rendered images. However, as the input face image may have occlusions (e.g., self-occlusion, glasses, and masks), we only compute the Euclidean distance between certain face regions $M_{face}$. The face regions are acquired by a pre-trained face segmentation networks following~\cite{shi2019face} on Halen dataset~\cite{le2012interactive}, and we use the regions of the face, eyebrows, eyes, nose, and mouth to compute the loss. Some examples are illustrated in Fig.~\ref{fig:skin_region}.

The pixel-wise loss is defined as:
\begin{equation}
   L_{pix} (x, x') = \frac{\sum M_{proj} M_{face} ||x - x'||_2}{\sum M_{proj} M_{face}},
\label{con:pixel_loss}
\end{equation}
where $x$ is the input image, $x'$ is the rendering image with illumination, and $M_{proj}$ denotes the region that the face mesh can be projected onto.

\subsubsection{Identity-Preserving Loss}
Using the pixel-level loss could generate a generally good result. However, the reconstructed 3D face might not "look like" the input 2D face image, especially under extreme circumstances. To tackle this issue, we define a loss function at the face feature level, which is called the identity-preserving loss. The loss requires the features between the input image and rendered image, extracted by the FaceNet, should be close to each other. We define the identity-preserving loss as a cosine distance:
\begin{equation}
   L_{id}(x, x') = 1 - \frac{<F(x), F(x')>}{||F(x)||\cdot||F(x')||},
\label{con:identity_loss}
\end{equation}
where $x$ is the input image, $x'$ is the rendered image, and $F(\cdot)$ is the feature extraction function by FaceNet. $<F(x), F(x')>$ is the inner product.

\subsubsection{Vertex-wise Loss}
When training the refinement module, the graph convolutional networks may not learn the texture RGB values properly from the image due to the occluded regions on the face. Therefore we construct a vertex-wise loss based on the texture predicted by the Regressor as assistance at the early stage of refinement module training, and then reduce the weights of the vertex-wise loss gradually. In order to obtain the small details from a human face, we also retrieve the vertex colors $T_p$ by projecting the face vertices to the 2D image, then feed them to the Refiner along with 3DMM texture RGB values. Taking those two into consideration, our vertex-wise loss can be defined as:
\begin{equation}
   L_{vert}(x, x') = \frac{1}{N} \sum_{i = 1}^{N} ||x_i - x'_i||_2,
\label{con:vertex_loss}
\end{equation}
where $N$ is the number of vertices. $x_i$ is the albedo value $T$ generated by the Regressor or the retrieved vertex color $T_p$, and $x'_i$ is the refined albedo $T'$ from the refinement module or the illuminated $T'$, denoted as $\widetilde{T'}$.

\subsubsection{Adversarial Loss}
For the adversarial training, we adopt the paradigm of Improved Wasserstein GAN~\cite{gulrajani2017improved}, whose adversarial loss is defined as:
\begin{equation}
\begin{aligned}
   L_{adv}(x, x') = &\mathop{\mathbb{E}}_{x' \sim \mathbb{P}_{R'}} [D(x')] -
   \mathop{\mathbb{E}}_{x \sim \mathbb{P}_I} [D(x)] +\\
   &\lambda \mathop{\mathbb{E}}_{\hat{x} \sim \mathbb{P}_{\hat{x}}} [(||\nabla_{\hat{x}} D(\hat{x})||_2 - 1)^2],
\end{aligned}
\label{con:adv_loss}
\end{equation}
where $x'$ is the rendered image from the refined texture and $x$ is the input image. $\hat{x}$ is the random sample which is uniformly sampled along the straight lines between the points sampled from the input image distribution $\mathbb{P}_I$ and the rendered image distribution $\mathbb{P}_{R'}$.

\section{Implementation Details}
Before training, every face image has been aligned following the method of~\cite{chen2016supervised}. Then we use the networks from~\cite{shi2019face} pre-trained on the Halen dataset~\cite{le2012interactive} for face segmentation, generating a face region mask for every face image. The face image dataset we use is CelebA~\cite{liu2015deep}, and the 3D morphable face model is Basel Face Model~\cite{paysan20093d}. The Regressor network is adopted from~\cite{deng2019accurate}, which predicts the 3DMM coefficients, pose and lighting parameters, and generates the face shape and coarse texture via the 3DMM model. We set the input image size to $224 \times 224$ and the number of vertices to 35,709, the same as~\cite{deng2019accurate}.

After getting the face shape and coarse texture, the next step is to train the texture refinement module and discriminator. The training loss is defined as:
\begin{equation}
\begin{aligned}
   L = &\sigma_1 [L_{pix}(I, R') + \sigma_2 L_{id}(I, R') + \sigma_3 L_{adv}(I, R')]\\
             &+ \sigma_4 [L_{vert}(T, T') + L_{vert}(T_p, \widetilde{T'})],
\end{aligned}
\label{con:loss}
\end{equation}
where $\sigma_2=0.2$ and $\sigma_3=0.001$ are fixed during training. As for $\sigma_1$ and $\sigma_4$, we train the texture refinement module starting with a "warm up" manner. Initially, we set $\sigma_1=0$ and $\sigma_4=1$. After one epoch of warm up, $\sigma_1$ is gradually increased to 1 and $\sigma_4$ is decreased to 0 accordingly.

During inference, our network can process about 50 images every second in parallel with an NVIDIA 1080Ti GPU.

\section{Experimental Results}
In this section, we demonstrate our results of the proposed framework and compare the reconstructed 3D faces with prior works. We also uploaded a video rotating the 3D face to YouTube\footnote[2]{https://youtu.be/ZGdkfhsd\_Hw} to show our results at large poses.

\begin{figure*}
\begin{center}
\includegraphics[width=0.74\linewidth]{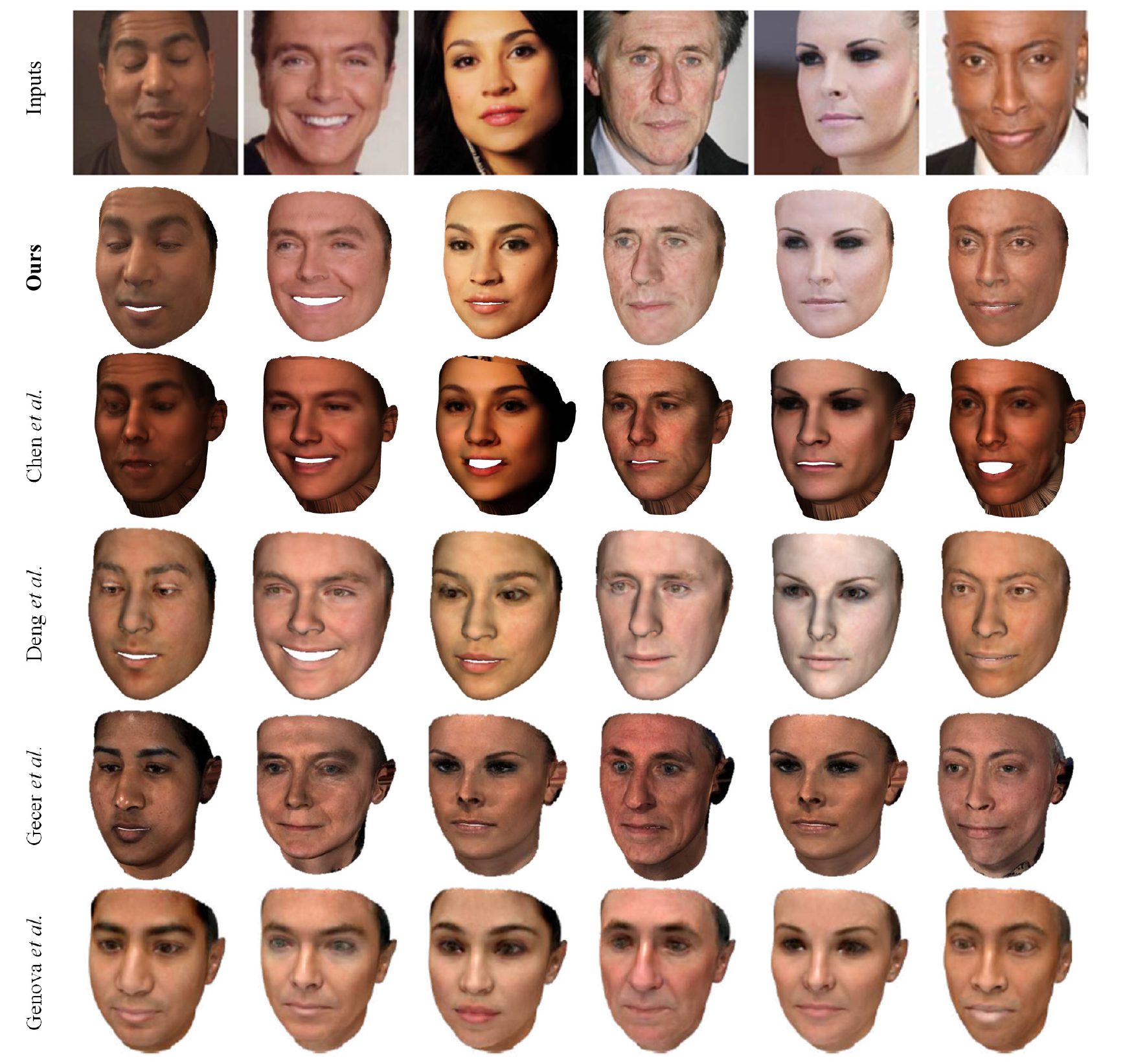}
\end{center}
   \caption{Comparison of our results with other methods. The first row is input images, our results are shown in the second row, and the remaining rows are reconstructed 3D faces obtained by~\cite{chen2019photo, deng2019accurate, gecer2019ganfit, genova2018unsupervised}, respectively.}
\label{fig:compare}
\end{figure*}

\subsection{Qualitative Comparison}
Fig.~\ref{fig:compare} shows our results compared to the most recent methods~\cite{chen2019photo, deng2019accurate, gecer2019ganfit, genova2018unsupervised}. The first row shows the input images and the second row shows our results. The remaining rows demonstrate the results of Deep Facial Details Nets~\cite{chen2019photo}, Deep 3D Face Reconstruction~\cite{deng2019accurate}, GANFIT~\cite{gecer2019ganfit}, and the method of Genova \etal~\cite{genova2018unsupervised} which introduced a differentiable renderer. 

It is worth to mention that in~\cite{chen2019photo, gecer2019ganfit}, they used their own captured data, which is not publicly available. Specifically, Chen \etal~\cite{chen2019photo} captured 366 high-quality 3D scans from 122 different subjects, and Gecer \etal~\cite{gecer2019ganfit} trained a progressive GAN to model the distribution of UV representations of 10,000 high-resolution textures, but we only make use of the in-the-wild 2D face image dataset.

The method of~\cite{chen2019photo} used UNets to produce displacement maps for facial details synthesis, and the networks are trained semi-supervised, with labeled 3D face exploited. However, their work mostly focused on local face shape details, the global shape of the reconstructed 3D face are not as good as the locals. 

With a large scale UV map dataset,~\cite{gecer2019ganfit} can generate high fidelity textures, but the results are lack of diversity in the respect of lights and holistic colors, as the predicted lights and colors are not so accurate.

Genova \etal~\cite{genova2018unsupervised} introduced an end-to-end, unsupervised training procedure, to learn to regress 3DMM coefficients. Deng \etal~\cite{deng2019accurate} predicted the lights and face pose simultaneously with 3DMM coefficients, which generated a better result under complex lighting situations. Still, our results have more detailed information since theirs are produced by a 3DMM model, which is not able to handle in-the-wild situations very well.

\subsection{Quantitative Comparison}
\begin{figure}[h]
\begin{center}
\includegraphics[width=0.8\linewidth]{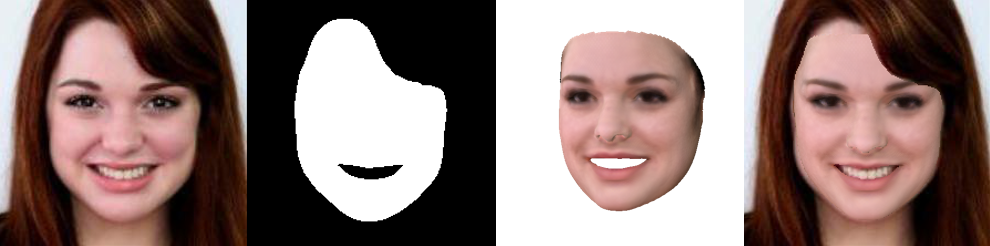}
\end{center}
   \caption{From left to right: input image, face region mask, reconstructed 3D face mesh, and the projected image for quantitative comparison.}
\label{fig:metric}
\end{figure}

To quantitatively evaluate our reconstructed 3D face, we employ the metrics on 2D images, since we do not have 3D ground-truth data. The metrics are computed between the input images and the projected images from face meshes, specifically, when the face mesh is projected to a 2D image, only the face region is projected, Fig.~\ref{fig:metric} shows an example.

Firstly, we use the $L_1$ distance, peak signal-to-noise ratio (PSNR) and structural similarity index (SSIM), which measure the results in pixel level. Then we evaluate the results in perception level by calculating the cosine similarity of feature vectors, which are extracted by two state-of-the-art pretrained face recognition networks, LightCNN~\cite{wu2018light} and evoLVe~\cite{zhao2019multi}. We do not use FaceNet because it appears in our training pipeline, making it unsuitable as an evaluation metric. Since most 3D face reconstruction methods focus on the shape of the face, not the texture, making it difficult to find enough metrics to compare with others. Additionally, we also compare the earth mover's distance between distributions of VGG-Face~\cite{parkhi2015deep} feature vector similarity from same and different identities on MICC datasets~\cite{Bagdanov:2011:FHF:2072572.2072597}. The quantitative comparison is shown in Tab.~\ref{tab:compare}.

Our results are better than others in multiple evaluation metrics. A lower $L_1$ distance and higher PSNR and SSIM indicate that our reconstructed 3D face textures is more close to the input images in pixel level. And both state-of-the-art face recognition networks believe that our results and the input images are more likely to be the same person.

In terms of MICC datasets, the facial meshes used in \cite{genova2018unsupervised} contain ears and neck region, while ours not. And we follow their procedure, only use the rendered images (third image in Fig.~\ref{fig:metric} demonstrates an example) for computing similarity score. The above factors may lead to lower scores of our method.

\vspace*{-4mm}

\subsection{Ablation Study}\label{sec:ablation}
\begin{figure}[h]
\begin{center}
\includegraphics[width=\linewidth]{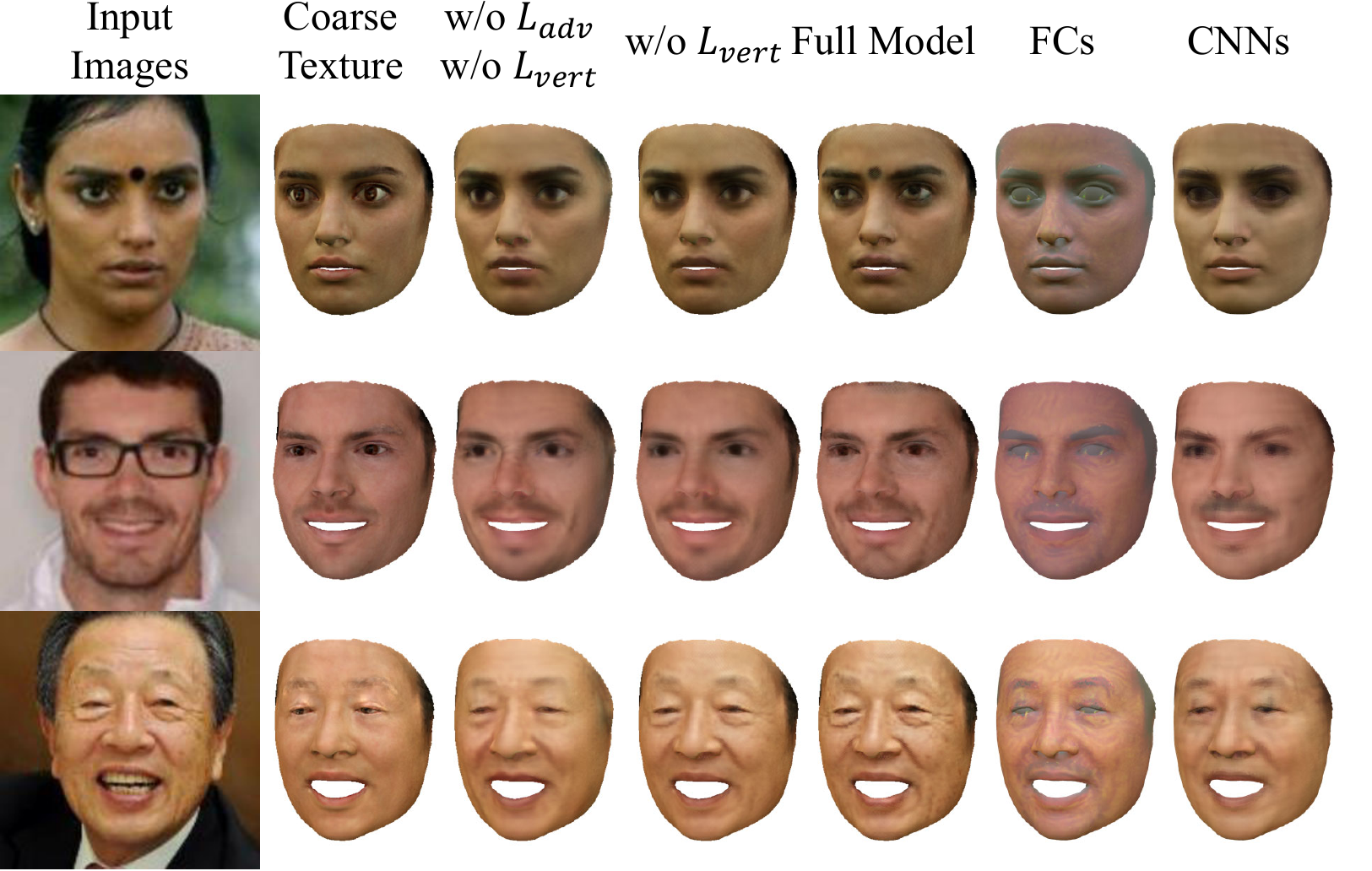}
\end{center}
   \caption{Ablation study on different losses of our proposed approach, the full model leads to a better result than others.}
\label{fig:ablation}
\end{figure}

Fig.~\ref{fig:ablation} demonstrates the ablation study of our approach, where our full model presents a more detailed and realistic texture than its variants. Tab.~\ref{tab:ablation} shows the quantitative metrics. The coarse texture is generated by the Regressor and 3DMM model, which is able to produce a basic shape and texture in general, however, the details such as lentigo or eyes are not faithfully predicted.



\begin{table}[t]
\begin{center}
\setlength{\tabcolsep}{3pt}{
\begin{tabular}{|c|c|ccc|c|}
\hline
&  & \cite{deng2019accurate} & \cite{genova2018unsupervised} & Ours & \cite{deng2018uv}* \\
\hline
\multirow{6}*{\rotatebox{90}{CelebA}}
& $L_1$ distance $\downarrow$ & 0.052 & /  & \textbf{0.034} & / \\
& PSNR $\uparrow$ & 26.58 & /  & \textbf{29.69} & 22.9$\sim$26.5 \\
& SSIM $\uparrow$ & 0.826 & /  & 0.894 & 0.887$\sim$\textbf{0.898} \\
& LightCNN $\uparrow$ & 0.724 & /  & \textbf{0.900} & / \\
& evoLVe $\uparrow$ & 0.641 & /  & \textbf{0.848} & / \\
\hline
\multirow{2}*{\rotatebox{90}{MICC}}
& \tabincell{c}{VGG-Face\\same} $\downarrow$ & 0.09 & 0.09 & \textbf{0.08} & / \\
& \tabincell{c}{VGG-Face\\diff.} $\uparrow$ & 0.11 & \textbf{0.32} & 0.11 & / \\
\hline
\end{tabular}}
\end{center}
\caption{Quantitative comparison. *: as a reference, we also list the PSNR and SSIM of~\cite{deng2018uv}, but it should be noted that their metrics are computed on UV maps since they have built a dataset containing 21,384 real UV maps for training, while we do not use such data.}
\label{tab:compare}
\end{table}

Starting by $L_{pix}$ and $L_{id}$, the networks are no longer restricted to a 3DMM model, and predicts higher fidelity face skin and eyes. With the help of adversarial training, the results become less blurry and more realistic. Finally, we construct our full model by adding the $L_{vert}$, the predictions contain more details, and looked very similar to the input image.

We also replace GCNs with fully-connected layers or convolutional layers on unwrapped UV space, and the performance are no better than GCNs, as shown in Fig.~\ref{fig:ablation} and Tab.~\ref{tab:ablation}. We reach to the similar conclusion as~\cite{zhou2019dense}, that using FCs or CNNs on UV spaces leads to a large number of parameters in the network and does not utilize the spatial information of the 3D facial structure.


\begin{table}[h]
\begin{center}
\setlength{\tabcolsep}{2pt}{
\begin{tabular}{|cccc|cccc|}
\hline
\multicolumn{4}{|c|}{Losses} & \multirow{2}*{PSNR} & \multirow{2}*{SSIM} & \multirow{2}*{LightCNN} & \multirow{2}*{evoLVe}\\
$L_{pix}$ & $L_{id}$ & $L_{adv}$ & $L_{vert}$ & & & & \\
\hline
 &  &  &  & 26.58 &0.826 & 0.724 &  0.641 \\
\checkmark & \checkmark & & & 28.57& 0.863 & 0.828 & 0.738 \\
\checkmark & \checkmark & \checkmark & & 29.30 & 0.872 & 0.840 & 0.755\\
\checkmark & \checkmark & \checkmark & \checkmark & \textbf{29.69} & \textbf{0.894} & \textbf{0.900} & \textbf{0.848}\\
\hline
\multicolumn{4}{|c|}{Fully-connected} & 25.88 & 0.820 & 0.629 & 0.544 \\
\multicolumn{4}{|c|}{Convolutional} & 27.54 & 0.848 & 0.798 & 0.696 \\
\hline
\end{tabular}}
\end{center}
\caption{Metrics on ablation study, higher is better.}
\label{tab:ablation}
\end{table}

\begin{figure}[h]
\begin{center}
\includegraphics[width=0.7\linewidth]{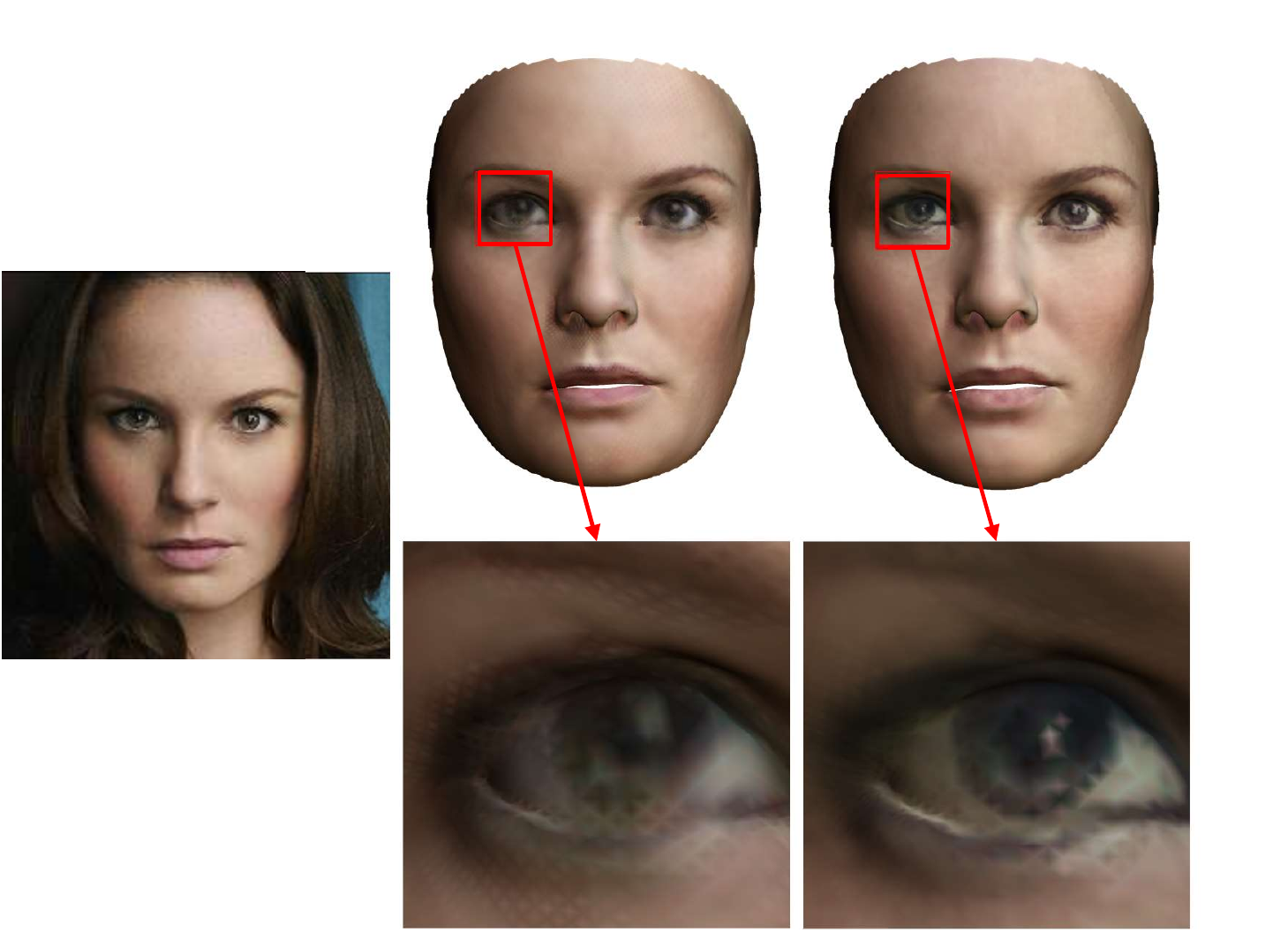}
\end{center}
   \caption{Comparison with higher resolution image. From left to right: the input image, result generated from resolution of $224 \times 224$ and $512 \times 512$.}
\label{fig:high_res}
\end{figure}

To improve the results, we train and test our proposed networks on a higher resolution image dataset, CelebA-HQ~\cite{karras2017progressive}, as well. Higher resolution helps to reduce the checkerboard-like artifacts and produce better results. A comparison is shown in Fig.~\ref{fig:high_res}.

\vspace*{-8mm}

\section{Conclusions}
In this paper, we present a novel 3D face reconstruction method, which produces face shapes with high fidelity textures from single-view images. To the best of our knowledge, we are the first to use graph convolutional networks to generate high fidelity face textures from the single images. Compared with other methods, our method does not require capturing high-resolution face texture datasets but can generate realistic face textures with just a 3DMM model. In the future, we will try to minimize the checkerboard-like artifacts and generate more detailed face shapes and expressions using graph convolutional networks as well.

\section*{Acknowledgments}
We thank anonymous reviewers for their valuable comments. The work was supported by National Key R\&D Program of China (2018YFB1004300), NSF China (No. 61772462, No. 61572429, No. U1736217) and the 100 Talents Program of Zhejiang University.

\clearpage

{\small
\bibliographystyle{ieee_fullname}
\bibliography{face_color}
}

\clearpage

\end{document}